\documentclass[letterpaper, 10 pt, conference]{ieeeconf}  

\IEEEoverridecommandlockouts                              

\usepackage{graphicx}
\graphicspath{ {./images/} }
\usepackage{caption}
\usepackage{subcaption}
\usepackage{amsmath}
\usepackage{amsfonts} 
\usepackage[svgnames]{xcolor}
\usepackage[hyperfootnotes=false,hidelinks]{hyperref}\usepackage{lipsum}
\usepackage{multicol}
\usepackage{graphicx}
\usepackage{soul}
\usepackage{cite}
\title{\LARGE \bf 
In-hand manipulation planning using human motion dictionary}

\author{Ali Hammoud$^{1}$, Valerio Belcamino$^{2}$, Alessandro Carfi$^{2}$,  Veronique Perdereau$^{1}$ and Fulvio Mastrogiovanni$^{2}$
\thanks{*This work was supported by the European project Index}
\thanks{$^{1}$A. Hammoud and V. Perdereau are with Sorbonne Universite, Institut des Systemes Intelligents et de Robotique,
ISIR, F-75005 Paris, France
        {\tt\small ali.hammoud, veronique.perdereau@sorbonne-universite.fr}}%
\thanks{$^{2}$V. Belcamino, A. Carfì and F. Mastrogiovanni are with TheEngineRoom,  Department of Informatics, Bioengineering, Robotics, and Systems Engineering, University of Genoa, Via Opera Pia 13, 16145, Genoa, Italy
        {\tt\small valerio.belcamino@edu.unige.it, alessandro.carfi@dibris.unige.it, fulvio.mastrogiovanni@unige.it.}}
}

\begin{document}

\maketitle
\thispagestyle{empty}
\pagestyle{empty}

\begin{abstract}
Dexterous in-hand manipulation is a peculiar and useful human skill. This ability requires the coordination of many senses and hand motion to adhere to many constraints. These constraints vary and can be influenced by the object characteristics or the specific application. One of the key elements for a robotic platform to implement reliable in-hand manipulation skills is to be able to integrate those constraints in their motion generations. These constraints can be implicitly modelled, learned through experience or human demonstrations. We propose a method based on motion primitives dictionaries to learn and reproduce in-hand manipulation skills. In particular, we focused on fingertip motions during the manipulation, and we defined an optimization process to combine motion primitives to reach specific fingertip configurations. The results of this work show that the proposed approach can generate manipulation motion coherent with the human one and that manipulation constraints are inherited even without an explicit formalization.
\end{abstract}
\section{INTRODUCTION}
Humans' daily activities often require interacting with objects and dexterously manipulating them in hand. These abilities result from a continuous learning process, during human lives, based on the observation of other humans' actions and personal attempts and errors. For robots to successfully integrate and operate in the human environment, they should interact with the environment as humans do \cite{carfi2021hand}. Therefore, robots should be able to manipulate unknown objects dexterously, adapting their previous experience to new scenarios. Furthermore, robots should be able to learn new manipulation skills by observing other "agents" actions. A robotic platform can achieve this by integrating advanced perception tools and flexible learning methods to represent and plan new manipulation actions.

Given a predefined manipulation goal, planning a dexterous manipulation consists in determining the necessary finger trajectories to reach it. Following the literature, we can divide approaches for planning robotic manipulations into two categories: data-driven and classical \cite{theodorou2010generalized, ijspeert2002learning, monahan1982state, prieur2012modeling, sundaralingam2017relaxed, murray1994mathematical, sundaralingam2018geometric, fan2017real, 9659421}. In data-driven approaches, dexterous manipulation models are trained either by robot trial and error or by observing human demonstrations \cite{theodorou2010generalized, ijspeert2002learning, monahan1982state, prieur2012modeling}. Instead, classical approaches are based on robotics principles, dividing complex tasks into sets of elementary actions \cite{sundaralingam2017relaxed, murray1994mathematical,sundaralingam2018geometric, fan2017real}. 

\begin{figure}[t]
  \centering
  \includegraphics[trim = 350 300 350 300, clip, width=0.48\textwidth]{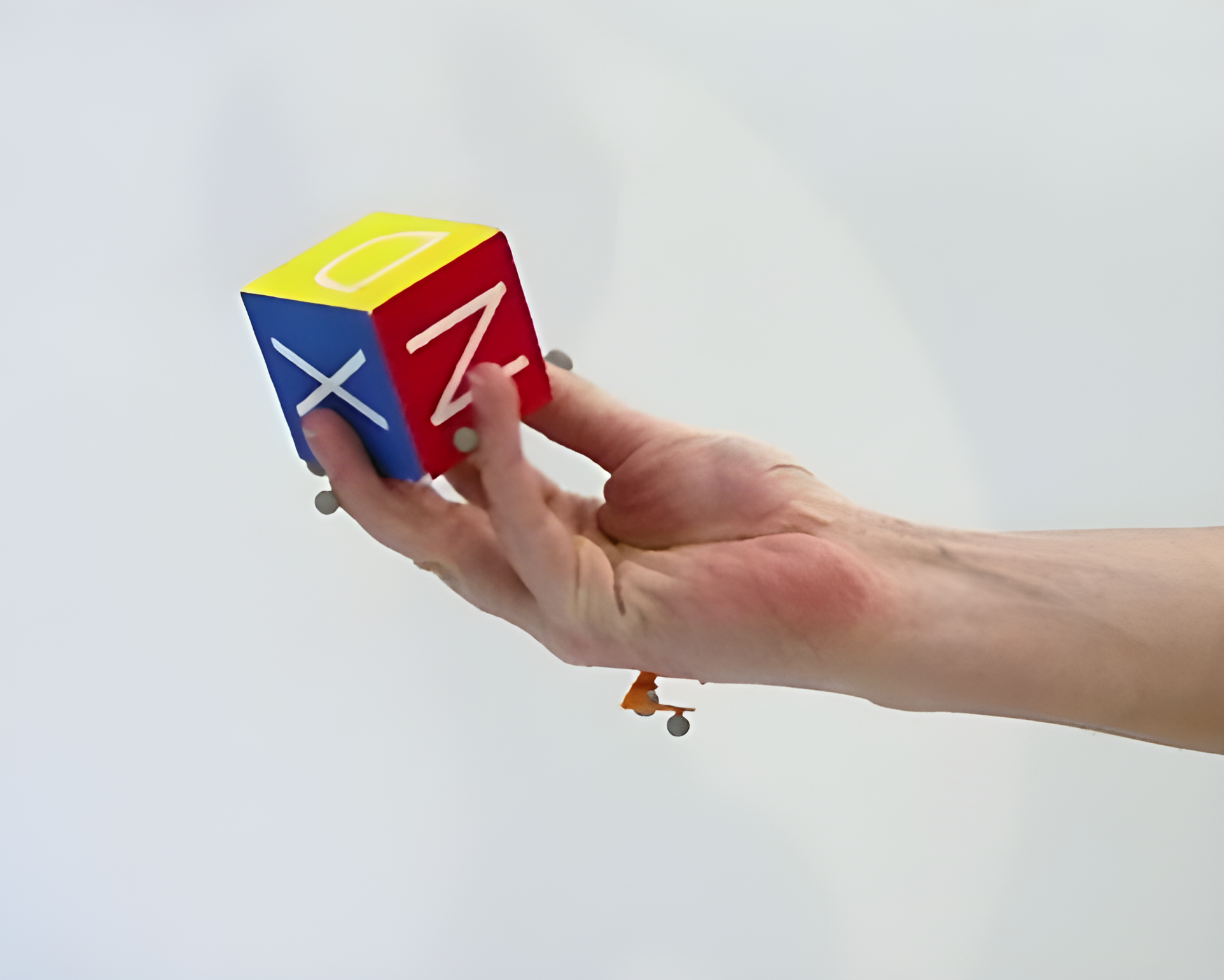}
  \caption{Example of an in-hand manipulation supported by the fingertips.}
  \label{fig:example}
\end{figure}

In the context of in-hand manipulation path planning, most data-driven approaches rely on dynamic movement primitives (DMP) \cite{theodorou2010generalized}. This solution is elegant and generates smooth trajectories while keeping a small number of parameters. DMP is made up of a set of generalized dynamics system equations that flexibly express movements. With DMP, it is possible to produce smooth movements of any shape by altering a simple linear dynamical system with a non-linear component \cite{ijspeert2002learning}. Non-linear components of DMP systems can be determined using data either from human demonstrations or robot ones.
Alternative systems use Markov Decision Processes (MDP). With MDP, a Dynamical Bayesian Network models the evolution of an agent's state according to its actions and the environment dynamics \cite{monahan1982state}. Finding an in-hand manipulation grasp sequence to pass from an initial hand position to a final one is what the MDP idea entails \cite{prieur2012modeling}. However, this does not include movement between grasps. In general, in-hand manipulation based on training methods necessitates a large amount of training data and a significant processing time.

On the other side, in-hand manipulation planning in classical robotics is solved by modelling the environment and the robotic hand. In this context, robot hand actions are decomposed into atomic sequences such as the in-hand regrasping and finger reallocation. In-hand regrasping consists in re-positioning the object to the desired pose within the robotic hand. Sundaralingam and Hermans 2017\cite{sundaralingam2017relaxed} proposed an efficient solution to this problem with a purely kinematic trajectory optimization. Various approaches based on rolling and sliding motions have also been explored\cite{murray1994mathematical}. Instead, in finger relocation, sometimes known as finger gaiting, the robot shifts contact points between the object and the fingertip by moving a single finger at a time. Solutions to finger gaiting based on geometric approaches were presented by Sundaralingam and Hermans, 2018 \cite{sundaralingam2018geometric} and Fan et al., 2017\cite{fan2017real}. Classical approaches have a predictable planning outcome since the designer sets all the constraints. However, this category of solutions can only approach atomic actions and not the whole in-hand manipulation due to the increasing complexity in modelling the system constraints.

Both classical and data-driven approaches have their limitations. With the former, one should manage large training datasets and long training times while, with the latter, one encounters difficulties in modelling complex constraints. We aim to overcome these limitations by generating human-like gaiting and in-hand regrasping with a hybrid approach that creates an in-hand manipulation primitives dictionary from human demonstrations \cite{9659421}. With this approach, combining primitives, we create robotic in-hand manipulation trajectories that respect the constraints imposed by the environment. This solution is possible since the dictionary is built from human demonstrations, and therefore, the dictionary elements respect all the complex constraints without the need for their formal definition. Finally, the possibility to build the primitives dictionary from small data samples resolves the computational problem often associated with data-driven approaches.

The remainder of this paper is organized as follows.
The concept of a primitive dictionary and its application into in-hand manipulation are covered in the first section. In the second section, path planning by integrating an in-hand manipulation dictionary is explained. The third part presents the implemented steps to integrate an in-hand manipulation dictionary into the path planning problem passing through data acquisition and model training and testing. In the fourth part, we show the obtained testing results for simulated trajectory .
\section{Primitives Dictionary}
\label{sec:formalization}
Primitives dictionaries have been extensively used in fields such as computer vision \cite{cao2020nmf}, clustering of documents \cite{akata2011non}, astronomy \cite{berne2007analysis} and generating human motion trajectories \cite{vollmer2014sparse,ijspeert2002movement,lewicki2000learning}. Given the desired behaviour to represent ($V$) it is possible to reconstruct it with a dictionary of primitives ($W$) combining them with a set of weights ($h$):
\begin{equation}
    V = W h
    \label{eq:primitives}
\end{equation}
We should point out that $W$ has dimensionality $|V|\times I$, where $|V|$ is the cardinality of the "behaviour to represent" and $I$ is the number of primitives. The weight vector $h$ has dimensionality $I$.
The primitives dictionary can be learned from data using different approaches. Most common solutions for primitive extractions involve the usage of non-negative matrix factorization method (NMF) \cite{sra2005generalized} or principal component analysis (PCA) \cite{hotelling1933analysis}. The selection process of the weights allows generating various behaviours starting from the same dictionary. For this reason, weights are typically selected through an optimization process considering constraints that can vary according to the scenario.

\subsection{In-Hand Manipulation Primitives}
In this work, we start from the assumption that it is possible to model the in-hand manipulation of an object (see Figure \ref{fig:example}), from an initial to the desired pose, as the evolution of the 3D position of the fingertips over time. The problem is therefore characterized by two main components: the fingers ($F(t)$) and object ($O(t)$) trajectories in space. Starting from this assumption, we can approach the problem by describing the fingertips as 3D points and their trajectories as follows:
\begin{equation*}
    F(t) = \left[ {\begin{array}{c}
    f_1(t)\\
    \vdots\\
    f_5(t) \\
  \end{array} } \right]\\
\end{equation*}
\begin{equation*}
    \textit{s.t.} \hspace{0.1 cm} t \in [1,\dots,N]
\end{equation*}
where 
\begin{equation*}
    f_i(t) = \left[ {\begin{array}{c}
    x_i(t)\\
    y_i(t)\\
    z_i(t) \\
  \end{array} } \right]\\
\end{equation*}
At the same time, we can describe the object pose evolution considering pose its 3D position and its orientation, expressed through Euler angles:
\begin{equation*}
    O(t) = \left[p_o(t), r_o(t) \right]\\
\end{equation*}
\begin{equation*}
    \textit{s.t.} \hspace{0.1 cm} t \in [1,\dots,N]
\end{equation*}
where
\begin{align*}
    & p_o(t) = \{x_o(t), y_o(t), z_o(t)\}\\
    & r_o(t) = \{\psi_o(t),\theta_o(t),\phi_o(t)\}
\end{align*}
Given these definitions the overall problem of the in-hand manipulation can be summarized as finding the intermediate points of the fingertips trajectories to carry the object from an initial pose ($O(1)$) to a final desired pose ($\hat{O}(N$) and reach a desired final fingertips configuration ($\hat{F}(N)$).

If we consider the problem one finger at a time and we recall Eq. \ref{eq:primitives} we can express the problem as follows:
\begin{equation}
    f_i(t) = W_i(t) h
    \label{eq:infusion}
\end{equation}
Where the primitives dictionary for the i-th finger is influenced by the time evolution. Given this formalization we can describe the overall problem as:
\begin{equation}
    F(t) = \mathbb{W}(t) h
    \label{eq:generation}
\end{equation}
where
\begin{equation*}
    \mathbb{W}(t) = \left[ {\begin{array}{c}
    W_1(t)\\
    \vdots\\
    W_5(t) \\
  \end{array} } \right]
\end{equation*}
Notice that with this description, the primitives are not described individually for each finger. Therefore, the weight $h$ is shared over all the finger primitives, and we describe the in-hand manipulation as the interplay of all the fingers.
With this formalization of the in-hand manipulation problem, it is possible to learn the primitives dictionary $\mathbb{W}$ using the non-negative matrix factorization method (NMF), as described by Hammoud et al., 2021 \cite{9659421}.

\section{Generation of Manipulations}
\label{sec:optimization}
Once the primitives dictionary is available, we can generate in-hand manipulation trajectories by infusing the dictionary with the weights. However, the generated manipulation should follow some specific constraints to be useful for robotic applications. Before introducing these constraints and explaining the procedure to generate new manipulation skills from our dictionary, we have to point out that the proposed approach is limited by a few assumptions:

\begin{itemize}
    \item The robot and gravity are the only factors that can affect the object's pose (i.e there are no external forces acting on our system).
    \item The robotic hand and the object are rigid.
    \item The initial object grasp is stable.
    \item The desired object pose is in the reachable zone of the fingertips.
    \item Contacts between the hand and the object are only made at fingertips.
    \item For fine in-hand manipulations, the final object pose is fully determined by the fingertips motions and the initial object pose.
\end{itemize}
Furthermore, we should point out that our formalization describes the in-hand manipulation as the trajectories in the Cartesian space of five fingertips, ignoring their orientations. This simplification is reasonable considering that, for in-hand manipulations, the fingertips' orientations, expressed with respect to the hand back, are strongly influenced by their 3D position. 
Although these assumptions limit the range of in-hand manipulation our method can achieve, we should point out that our method is able to produce a wide range of manipulations and that it could be extended to relax some of these assumptions. 
Given the problem formalization and the initial assumption, we can now introduce the procedure to compute the appropriate weights to solve a specific manipulation. This is done through an optimization process that has an objective function $J$ described as:
\begin{equation}
    J(F,O) = J_F + \alpha J_O
    \label{eq:optimization}
\end{equation}
where $J_F$ represents the distance between the final fingertips configuration $F(N)$ and the desired final fingertips configuration $\hat{F}(N)$. $J_F$ is defined as follows:
\begin{equation*}
    J_F = \sum^5_{i=1} \lVert f_i(N) - \hat{f}_i(N)\rVert
\end{equation*}
$J_O$ is the distance between the final object pose $O(N)$ and the desired final pose ($\hat{O}(N)$). Since the object pose is composed of both 3D position and orientation, the distances for the two components are computed separately and the overall distance $J_O$ is defined as the sum of the two distances:
\begin{equation*}
    J_O = \lVert p_o(N) - \hat{p}_o(N)\rVert + \lVert r_o(N) - \hat{r}_o(N)\rVert 
\end{equation*}
Finally, alpha is a scalar weight fine-tuning the trade-off between the two cost components.
While solving the in-hand manipulation problem for a robotic hand, other constraints linked to the robotic hand kinematics should be addressed. The i-th fingertip should be in the reachable workspace ($\mathbb{F}_i$)
\begin{equation}
    f_i(t) \in \mathbb{F}_i, \forall i \in [1,5], \forall t \in [1,N]
    \label{eq:reachability}
\end{equation}
and at the same time, each fingertip should respect kinematic constraints on their instantaneous velocity
\begin{equation}
    \dot{f}_{i,min} \leq \dot{f}_i(t) \leq \dot{f}_{i,max}, \forall i \in [1,5], \forall t \in [1,N]
    \label{eq:velocity}
\end{equation}
While performing the in-hand manipulation, other constraints apply. Fingertips should not collide with each other and at each instant, during the manipulation, the object should be in contact with at least three fingertips to ensure a stable grasp \cite{Prattichizzo2008}. We can express the first constraint by checking the Euclidean distances between fingertips for all the manipulation time:
\begin{equation}
 \min_{\forall j \in (i,5]} \lVert f_i(t) - f_j(t) \rVert \neq 0, \forall i \in [1,4], \forall t \in [0,N]
 \label{eq:collisions}
\end{equation}
For the second constraint, we have instead to introduce the distance between the fingertip ($f_i(t)$) and the object surface, represented as a point-cloud ($M(t)$) of 3D points ($m_k(t)$, with $k \in [1,|M|]$). Given this representation, the distance between the i-th fingertip and the object ($d_i(t)$) is defined as the distance between the fingertip and the closest point of the object point-cloud:
\begin{equation*}
    d_i(t) = \min_{\forall k \in [1,N]}\lVert f_i(t) - m_k(t) \rVert
\end{equation*}
and the subset of fingertips in contact with the object ($Fc(t)$)
\begin{equation}
    Fc(t) = \{ f_j(t) \in F(t) \hspace{0.1cm} | \hspace{0.1cm} f_j(t) \leq \tau\}
    \label{eq:single-contact}
\end{equation}
where $\tau$ is a threshold value defining the maximum distance between the fingertip and the object when they are in contact. Given this definition, the cardinality of the subset of fingertips in contact should be always at least 3 
to maintain a stable grasp during the manipulation
\begin{equation}
    |Fc(t)| \geq 3, \hspace{0.1 cm} \forall t \in [1,N]
    \label{eq:contacts}
\end{equation}
\subsection{Relaxed Optimization Problem}
Human manipulation typically follows the constraints previously described and our approach uses human data to derive the manipulation primitives. Therefore, it is reasonable to assume that at least some of these constraints are going to be transposed to the new generated trajectories without including them in the optimization problem. This consideration should apply for most of the constraints except for the fingertips velocity constraint because of the differences between the human and robot's velocity ranges. Furthermore, since each dictionary is trained for a specific object and the pose of the object is strictly related to the fingertips motions, because of the first assumption, we can simplify the optimization function considering the fingertips component only. Therefore, Eq. \ref{eq:optimization} can be rewritten as:
\begin{equation*}
    J(F) = J_F = \sum^5_{i=1} {\lVert f_i(N) - \hat{f}_i(N)\rVert}_{2}^{2}
\end{equation*}
that using Eq. \ref{eq:infusion} can be written as:
\begin{equation}
    J(F) = \sum^5_{i=1} {\lVert W_i(N) h - \hat{f}_i(N)\rVert}_{2}^{2}
    \label{eq:soptimization}
\end{equation}
The optimization process should find the weights $h$ that minimizes Eq. \ref{eq:soptimization} while considering the fingertips velocity constraint expressed in Eq. \ref{eq:velocity}.

All the other constraints not directly considered in the optimization phase (Eq. \ref{eq:reachability},\ref{eq:collisions},\ref{eq:contacts}) are going to be evaluated during the testing phase to determine if they are embedded into dictionaries learned from human demonstrations.
\begin{figure*}
    \centering
    \vspace{1cm}
    \begin{subfigure}[b]{0.49 \textwidth}
        \centering
        \includegraphics[height = 0.8\textwidth]{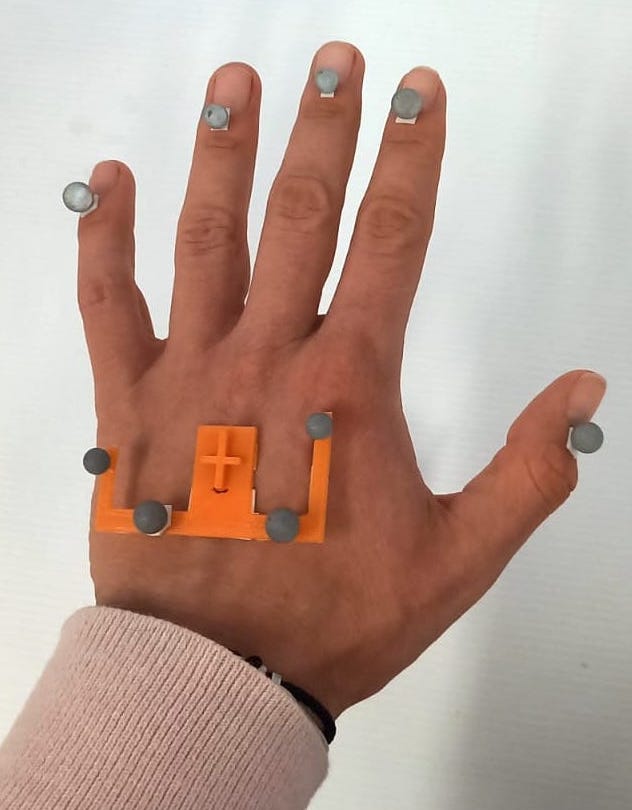}
        \caption{Hand Marker Placement.}
        \label{fig:hand}
    \end{subfigure}
    \begin{subfigure}[b]{0.49 \textwidth}
        \centering
        \includegraphics[trim = 160 0 200 30, clip, height= 0.8\textwidth]{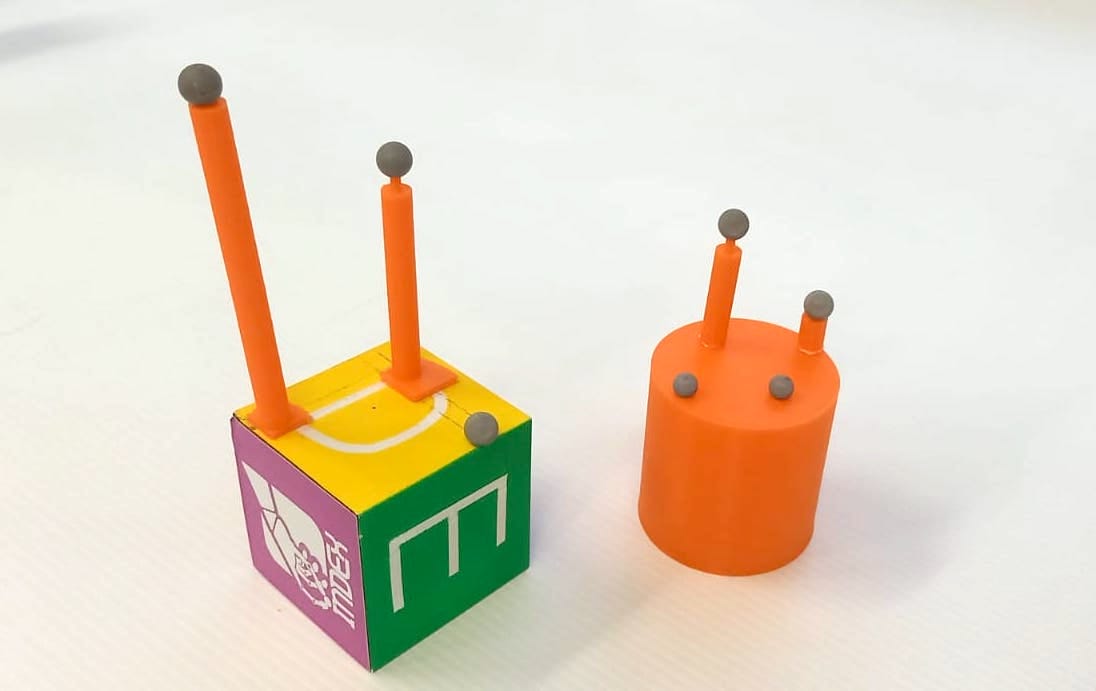}
        \caption{Objects Marker Placement.}
        \label{fig:objects}
    \end{subfigure}
    \caption{On the left the placement of the MoCap markers on the hand, on the right the placement of markers on the two different objects.}
\end{figure*}

\begin{figure}
  \centering
  \includegraphics[width=0.45\textwidth]{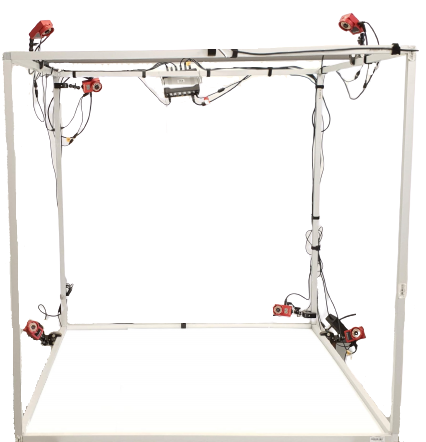}
  \caption{The motion capture system.}
  \label{fig:setup}
\end{figure}

\section{Implementation}
Given the problem statement introduced in Section \ref{sec:formalization} and the optimization procedure detailed in Section \ref{sec:optimization}, human demonstrations of in-hand manipulations are needed to implement and evaluate the presented approach. These demonstrations should consist of the 3D positions of the fingertips during manipulation. In fact, this information is enough to both train the dictionary and generate new manipulations. However, to test the adherence to the constraints introduced in Section \ref{sec:optimization}, the pose of the object is also necessary.

Many solutions exist to track the position and the orientation of objects. Information from RGB cameras can be processed to extract the position of the fingertips \cite{9659421} or datagloves can be used to track human motions \cite{gloveREF}. At the same time, the pose of an object can be tracked using embedded inertial measurement units \cite{niewiadomski2021multimodal} or through image processing \cite{park2019pix2pose}. Since the main objective of this work is to validate the proposed approach to generate in-hand manipulations, we used a motion capture (MoCap) system to gather precise and reliable data.

\subsection{Data Acquisition}
For the data acquisition we used a MoCap system composed of six OptiTrack Flex-3 cameras\footnote{\url{https://optitrack.com/cameras/flex-3/}} (see Figure \ref{fig:setup}). This MoCap system can track the 3D position of reflective markers (see Figure \ref{fig:hand} and \ref{fig:objects}), and position and orientation of sets of markers mounted on a rigid object, i.e., rigid body. The MoCap acquires all the data at the constant frequency of 100Hz.

Data about the hand motion during the manipulation are collected using five markers on the fingertips and one rigid body on the hand back (see Figure \ref{fig:hand}). More precisely, the reflective markers are placed between the nail and the distal-interphalangeal joint to minimize collisions and occlusions. In this way, we maximize the accuracy of the tracking system. The objects poses are tracked with a single rigid body. As can be seen in Figure \ref{fig:objects}, the experimental scenario involves two different objects: i) a cube with an edge of a 5 cm; ii) a cylinder with 5 cm diameter and 5 cm height.

For both objects, we have collected two datasets, one for training and another one for testing. For what regards the training dataset only the hand was tracked by the MoCap system and the experiments consisted of rotations and translations of the object on multiple axes while its weight is supported by the fingers. The palm of the hand keeps facing upwards during these actions for a stronger grip and to guarantee that the object is exclusively rotated by the movement of the fingers. Figure \ref{fig:example} shows a snapshot of the experiment.

To record the data for testing the proposed method, we adopted a slightly different approach since also tracking the object pose was necessary. As shown in Figure \ref{fig:objects}, a set of markers were attached to the objects to track their pose. The tracking of the object with the MoCap system inserts a trade off between the freedom of movement and the tracking precision. We opted to place the markers on a single face in a way that their centroid coincides with the axis of symmetry of the object. This choice allows precise tracking of the rigid body pose and easy computation of the absolute object position from the measurements with a simple translation in the local system of reference. However, this marker disposal limits the possible manipulation, allowing complete rotation along the axis orthogonal to the markers only.

Both training and testing data have been recorded thanks to a single subject performing for approximately 2 minutes an in-hand manipulation: five times for the training and one for the test. This procedure has been repeated for both objects.

\subsection{Model Training and Testing}
Before the training process can take place the collected data should be processed. The processing is divided into three phases. In the first phase, the position of the fingertips and the full pose of the object are transformed according to the hand back pose. Following the first assumption presented in Section \ref{sec:optimization}, we are going to study the components of the manipulation related to the fingertips only. Therefore, by performing the described transformation we can work only on the fingertip and object poses assuming the hand is static. In the second phase, all the processed data are filtered using a median filter of window size 50 samples. In the third and last phase, given the object model and its pose, it is generated a pointcloud ($M(T)$) of the object surface that will be used in the testing phase to compute the distance between the fingertips and the object. The final cube pointcloud has a dimension of around fifty thousand points while the cylinder one has a dimension of seventy thousand points.

After their processing the training data are used to derive two dictionaries, one for the cube and one for the cylinder, using the approach proposed by Hammoud et al., 2021 \cite{9659421}. Both dictionaries have been trained on 1 second segments of the full in-hand manipulation. This allows to simplify the training and to model intermediate configurations of the in-hand manipulation. However, this implies that these dictionaries could be used to generate in-hand manipulation of 1 second only. The final dictionaries have both 200 primitives (i.e., $I=200$).

Finally, to test the method, the collected data are divided into segments of 1 second. For each segment, the last fingertip configuration is used as target ($\hat{f}_i(N)$) for the optimization process described by Eq. \ref{eq:soptimization}. To solve the optimization problem and find the weights ($h$) we used the MATLAB optimization toolbox with a quadratic programming feature. The resulting weights are then used in Eq. \ref{eq:generation} to generate all the entire fingertips trajectory $F(t)$. Notice that since the trajectory length is 1 second and the data frequency is 100Hz, the trajectory is composed of 100 samples ($t \in [1,100]$). From the single sequence recorded for the testing, we were able to extract and generate 100 sequences of 1 second for each object.

\begin{figure}[t]
  \centering
  \includegraphics[width=0.48\textwidth]{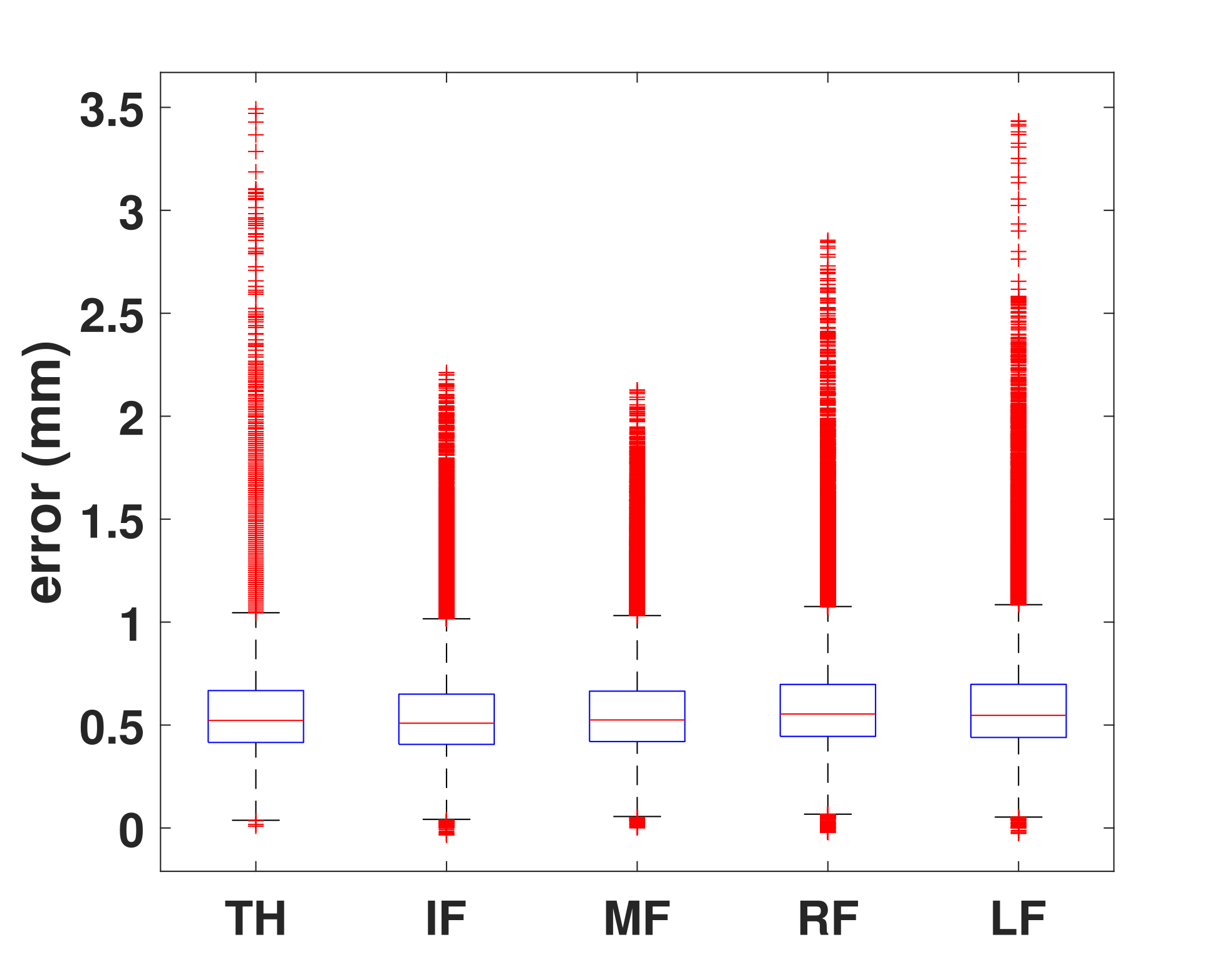}
  \caption{The box plot of the Euclidean distance between recorded and generated fingertips positions for the cube manipulation. TH - Thumb; IF - Index; MF - Middle; RF - Ring; LF - Little.}
  \label{fig:cube-error}
\end{figure}

\begin{figure}[t]
  \centering
  \includegraphics[width=0.48\textwidth]{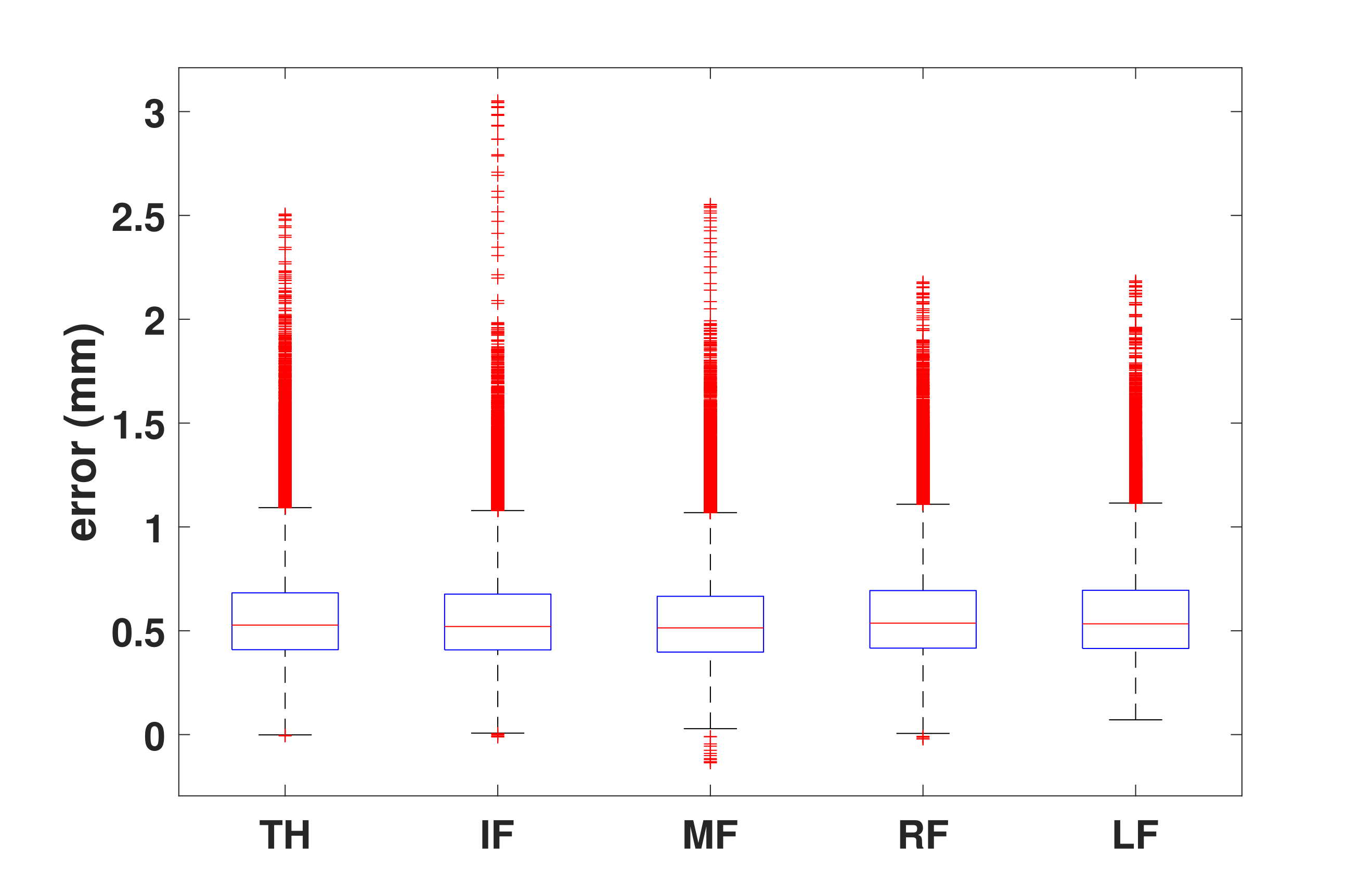}
  \caption{The box plot of the Euclidean distances between recorded and generated fingertips positions for the cylinder manipulation.}
  \label{fig:cylinder-error}
\end{figure}

\section{Results}
To evaluate the performance of our approach to generate new manipulation trajectories, we carried out an in-depth analysis. At first, we analyzed the error between recorded and generated manipulations. We computed this error at each instant as the Euclidean distance between recorded and generated points. These results are summarized by the box plot in Figure \ref{fig:cube-error} for the cube and Figure \ref{fig:cylinder-error} for the cylinder. These two figures show the errors for each finger over approximately ten thousand time instants. The results present many outliers, up to $3.5 mm$ of error. However, for all the fingers, in $75\%$ of the time instants, the error is below $1.2 mm$.

To check the capability of the system to generate new manipulation trajectories while preserving the characteristics of the human manipulation we tested the adherence of the new trajectories to the three constraints introduced in Section \ref{sec:optimization}: i) reachability, ii) finger collision, and iii) minimum contact points.

To test the reachability constraint we defined the reachable workspace $\mathbb{F}_i$, for each finger, as the set containing all the fingertip positions observed in the training set. Then we checked if the new generated fingertip positions are inside the reachable workspace as described in Eq. \ref{eq:reachability}. All the points generated with our approach satisfied the reachability constraint.

For the finger collision, we computed the minimum finger to finger distance as defined in Eq. \ref{eq:collisions}. Over the one-hundred trajectories generated for each object, we detected 4 time instants with a collision for the cube 
and only one for the cylinder. 
Therefore, the generated trajectories also satisfied this constraint.

\begin{figure}[t]
  \includegraphics[width=0.48\textwidth]{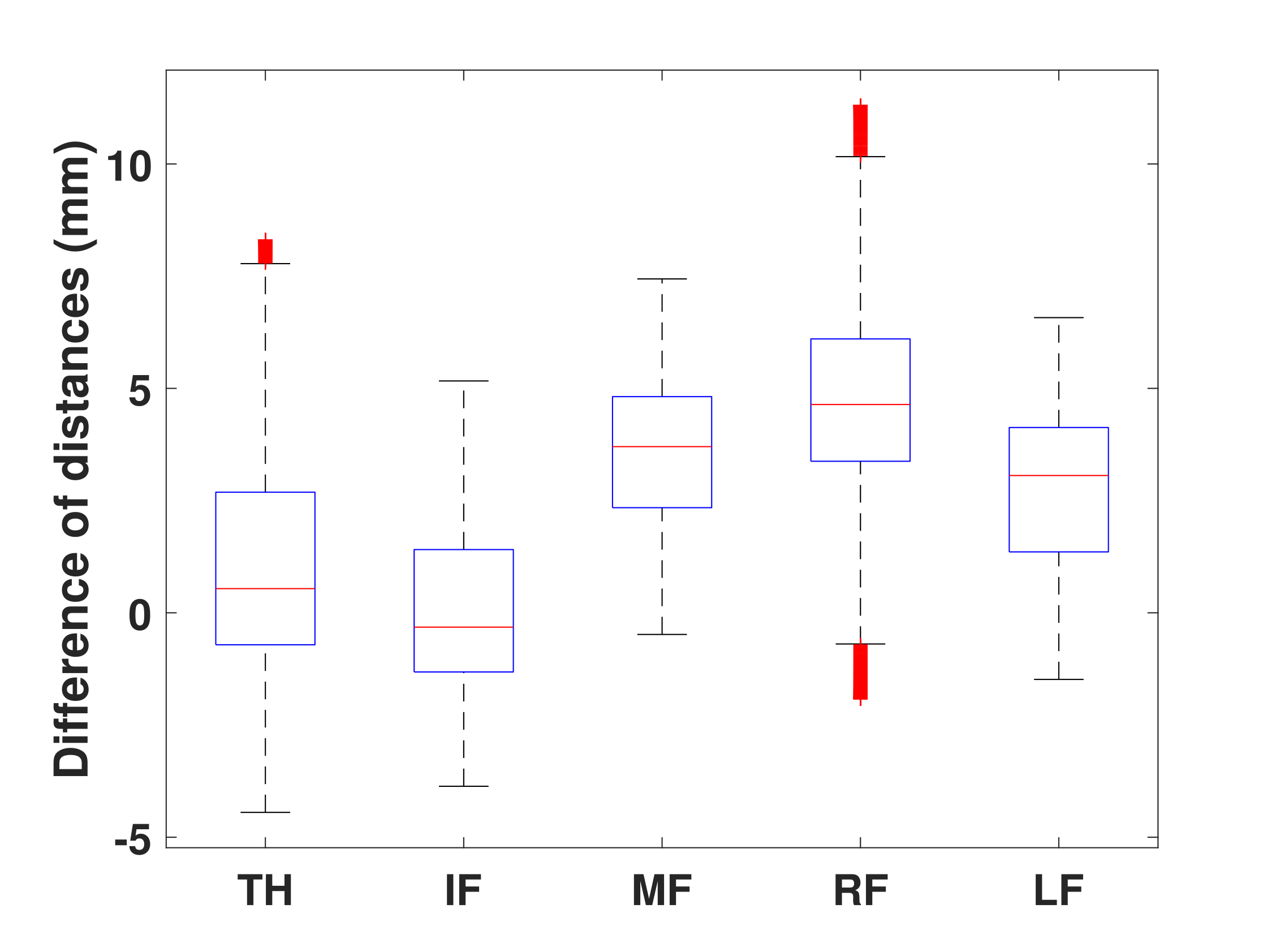}
  \caption{Difference of distances between human and object and created path and object for the testing cube data.}
  \label{fig:distance-cube}
\end{figure}

\begin{figure}[t]
  \includegraphics[width=0.48\textwidth]{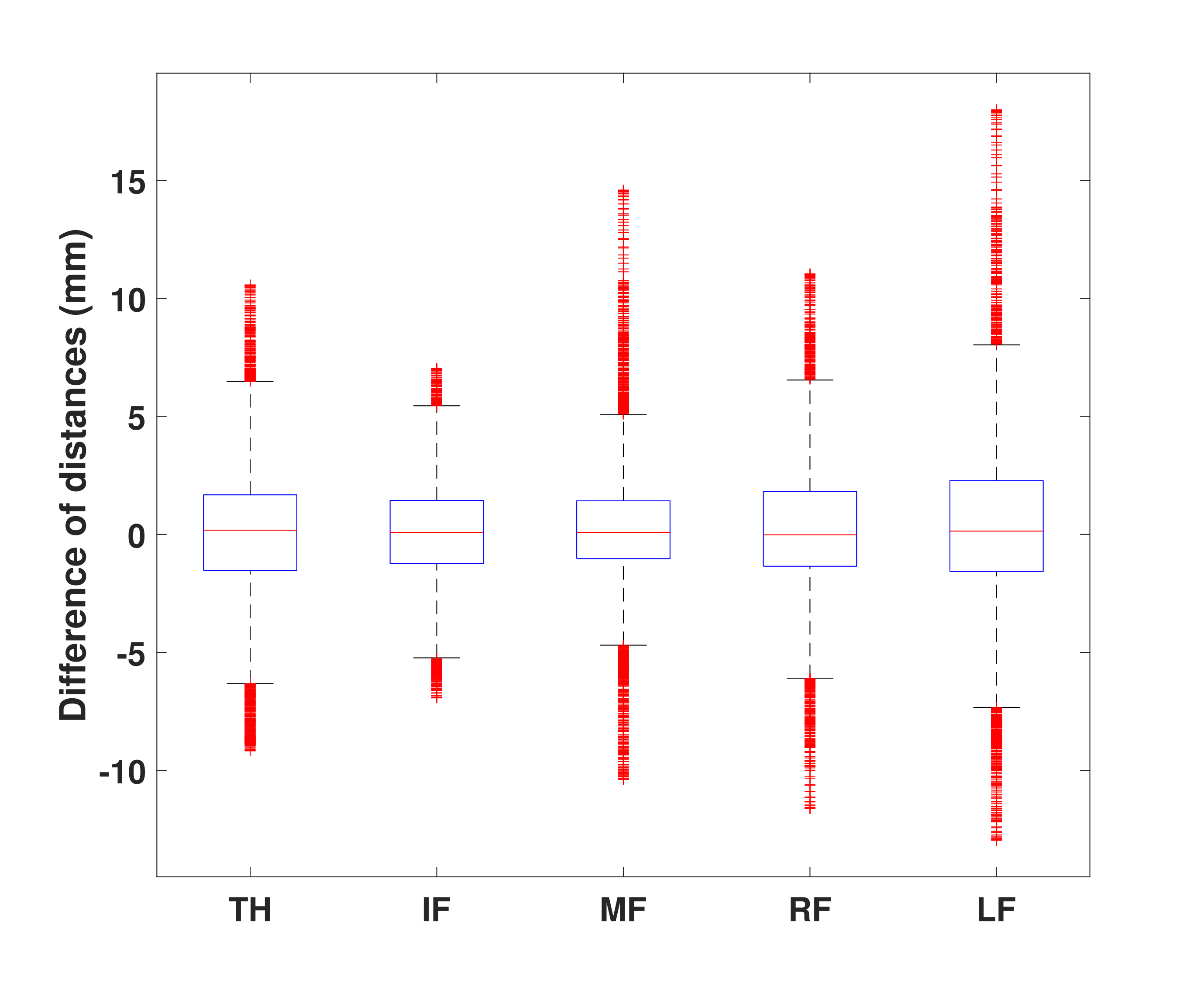}
  \caption{Difference of distances between human and object and created path and object for the testing cylinder data.}
  \label{fig:distance-cylinder}
\end{figure}

Finally, for what regards the minimum contact points between the fingertips and the object we decided to divide its evaluation into three steps: i) compute the finger to object distance both for recorded and generated data; ii) compute the distance difference between the recorded and generated data to check for similarities in the behaviours; and iii) given the finger to object distance for the generated data 
determine at each instant how many contacts are present. The Euclidean distance between the fingertips (recorded and generated) and the object has been computed as described in Section \ref{sec:optimization} using the "object pointcloud" generated during the data preprocessing. For each time instant of the one-hundred sequences, the difference between these two fingers to object distances is presented in the box plots of Figure \ref{fig:distance-cube} (for the cube) and \ref{fig:distance-cylinder} (for the cylinder). From the box plots we can see that, contrary to what happens for the cylinder, the difference varies a lot over the different fingers for the cube. This behaviour could be linked to the different shapes but we do not have enough details to conclude on it. Overall the median value for each finger over all the objects is between $0$ and $5 mm$ and maximum differences never overcome $15 mm$. According to this, we can conclude that the relation between the object and fingertips can be partially different when considering human and generated trajectories. This aspect should be the subject of future studies but does not imply that the minimum contact points constraints are not satisfied. To determine when a fingertip is in contact with the object we refer to Eq. \ref{eq:single-contact} setting $\tau = 5 mm$. This value has been chosen since according to Delhaye et al., 2021 \cite{delhaye2021high} the fingertip deformation can go up to $5 mm$. According to this parameter, the generated trajectories were able to maintain three points of contact for $97\%$ and $93\%$ of the time respectively for the cube and the cylinder.
 
\section{Conclusions}
This paper introduced a method based on motion primitives dictionaries to generate human-like fingertip trajectories for in-hand manipulation. The approach leverages human demonstrations to learn a manipulation dictionary that implicitly includes in-hand manipulation constraints.

The proposed approach, tested for the in-hand manipulation of two different objects (i.e., a cube and a cylinder), could generate new manipulation trajectories coherent with the human demonstrations. This coherence is also evident in the adherence of the manipulations with three constraints that characterizes the in-hand manipulation (i.e., reachability, finger collision, and minimum contact points). A relevant result is that the proposed approach inherits these constraints from the human demonstrations without the need for a formal representation.

Given the observed results, trajectories generated with this approach could drive robotic hands in dexterous manipulations. However, future work in this field should focus on relaxing the adopted assumptions to target more complex in-hand manipulations, for example, by extending the effect on the manipulation from the fingertips to other hand parts.

\addtolength{\textheight}{-12cm}   




\section*{ACKNOWLEDGMENT}
This work is supported by the CHIST-ERA (2014-2020) project InDex and received funding from Agence Nationale de la Recherche (ANR) under grant agreement No. ANR-18-CHR3-0004 and the Italian Ministry of Education and Research (MIUR).
\bibliography{mybib}{}
\bibliographystyle{ieeetr}

\end{document}